\documentclass[letterpaper, 10 pt, conference]{ieeeconf}  

\IEEEoverridecommandlockouts                              

\usepackage{graphics} 
\usepackage{epsfig} 
\usepackage{mathptmx} 
\usepackage{times} 
\usepackage{amsmath}
\usepackage{xcolor}
\usepackage{subcaption}
\usepackage{amssymb}
\usepackage{subcaption}
\usepackage{multicol}
\usepackage{hyperref}
\usepackage[amssymb]{SIunits}

\newtheorem{assumption}{Assumption}

\newtheorem{definition}{Definition}
\newtheorem{remark}{Remark}

\usepackage{cite}

\usepackage{units}

\makeatletter
\newcommand*\titleheader[1]{\gdef\@titleheader{#1}}
\AtBeginDocument{%
  \let\st@red@title\@title
  \def\@title{%
    \bgroup\normalfont\large\centering\@titleheader\par\egroup
    \vskip1.5em\st@red@title}
}
\makeatother

\title{\LARGE \bf
A CBF-Adaptive Control Architecture for Visual Navigation for UAV in the Presence of Uncertainties}
\titleheader{Preprint, 2024 IEEE International Conference on Robotics and Automation (ICRA 2024), Accepted}

\author{Viswa Narayanan Sankaranarayanan$^{*}$, Akshit Saradagi, Sumeet Satpute, and George Nikolakopoulos
\thanks{*Corresponding author}
\thanks{All the authors are with Robotics and Artificial Intelligence Group of the Department of Computer Science, Electrical and Space Engineering at Lule\aa \ University of Technology, Sweden. This work has been partially funded by the European Unions Horizon 2020 Research and Innovation Programme AERO-TRAIN under the Grant Agreement No. 953454.}}
\begin{document}
\maketitle
\thispagestyle{empty}
\pagestyle{empty}

\begin{abstract}
In this article, we propose a control solution for the safe transfer of a quadrotor UAV between two surface robots positioning itself only using the visual features on the surface robots, which enforces safety constraints for precise landing and visual locking, in the presence of modeling uncertainties and external disturbances. The controller handles the ascending and descending phases of the navigation using a visual locking control barrier function (VCBF) and a parametrizable switching descending CBF (DCBF) respectively, eliminating the need for an external planner. The control scheme has a backstepping approach for the position controller with the CBF filter acting on the position kinematics to produce a filtered virtual velocity control input, which is tracked by an adaptive controller to overcome modeling uncertainties and external disturbances. The experimental validation is carried out with a UAV that navigates from the base to the target using an RGB camera.
\end{abstract}

\section{Introduction} \label{sec:intro}
The field of multi-model robotics, which involves coordinated aerial and surface vehicles, is of research interest in many applications owing to their combined advantages. In such a scenario, the navigation of a quadrotor UAV from a surface robot (base) to another (target) presents interesting challenges to the robotic community \cite{perez2018architecture, wang2014efficient}. Further, it opens a gateway to applications such as payload transportation \cite{sankaranarayanan2020aerial} and aerial manipulation \cite{wuthier2017geometric}.

Along this direction, while many of the existing works in the literature focus on the landing platform design, target identification, and visual-servoing for landing \cite{grlj2022decade}, a systematic approach for navigation between the two objects is not well addressed in this domain. Further, a nominal navigation solution using planning and guidance \cite{kim2019uav, narvaez2020autonomous} requires precise knowledge of the poses of the UAV and target in the global frame. In a typical GNSS-denied environment, the UAV is equipped with a camera for odometry and navigation \cite{gyagenda2022review}. However, in a dynamic environment (or environment with reflective surfaces), the localization algorithms are too heavy for the onboard computer. This work considers such a scenario, where the relative localization is performed using the visual features on the surface robots, but a global localization is unavailable.

The control problem in this scenario involves two phases: ascending and descending. The existing control solutions based on onboard vision sensors \cite{lee2016vision, meng2019visual, recalde2022constrained, zhao2021robustVisual, lin2022robust, lin2021low, demirhan2020development, mu2023vision} broadly address the descending phase, which is referred to as the landing problem. While visual servoing detects and lands on the surface robot based on feature detection and tracking, an external control layer is required to enforce safety constraints since the UAV is prone to collision with the target, undesirable ground effects, loss of field of view (FOV), and landing precision based on the application and design of the landing platform. While some of the works handle the FOV constraints, there is no explicit control strategy to enforce precise landing (required for reducing the ground effects) and collision avoidance with the target, simultaneously. Further, the existing constrained control frameworks are prone to modeling uncertainties and external disturbances, such as wind. Besides, since these controllers handle only the landing phase, a control strategy has to be formulated to accumulate both the ascending and descending phases. In the absence of global localization, it is practical to know the approximate direction of the target with respect to the base though the exact landing spot is unknown.

Given these observations, a complete control framework for the navigation of a UAV between two surface robots exploiting the vision-based localization using the features on the surface robots that can handle landing precision and other safety constraints, and modeling uncertainties is still missing in the literature. Towards this problem, the following contributions are presented in this work:

\textbf{Contributions:} In this article, we present a switched control barrier function (CBF) based novel control approach to navigate a UAV between a base and a target, with only an a priori knowledge of the potential direction of the target, without any global localization. The ascending phase is handled by a visual-locking CBF (VCBF), where the UAV moves in the potential direction of the target while maintaining the features on the base in its field of view. The descending phase is converted into two sub-phases: approaching, and landing. When the target is detected, the controller switches to the approaching phase, where the UAV aligns itself over the target. Then, the controller switches to the landing phase where the UAV descends vertically to land. The sub-phases of descending are constrained using a single parameterizable descending CBF (DCBF), which switches its parameters based on the control phase and the relative position of the UAV. Finally, an adaptive controller is used at the lower level to track the desired velocity inputs generated by the CBF to compensate for the external disturbances. The controller is validated with an experimental scenario.

The rest of the article is organized as it follows: Section \ref{sec:prob_form} presents the model of the UAV and formulates the problem; Sections \ref{sec:con_arc} \& \ref{sec:cbf_design} explains the proposed control architecture and the CBF design respectively, which is validated in with experiments in Section \ref{sec:exp_val}; Section \ref{sec:conc} provides the concluding remarks and discusses the problems for future work.

\textbf{Notations:} The following notations are used in this article: $\mathbf{R}_A^B, \mathbf{T}_A^B$ respectively represent the rotation and transformation matrices of frame $B$ with respect to frame $A$; $\mathbf{p}_A^B$ is the position vector to the origin of frame $B$ expressed in frame $A$; $\mathbf{p}_A$ refers to the position of UAV in frame $A$; $||.||$ denotes the Euclidean norm; $L_f h(x)$ represents the Lie derivative of a continuously differentiable function $h(x)$ along the vector field $f(x)$, i.e., $\frac{\partial h(x)}{\partial x} f(x)$; the boundary of a set $\mathcal{S}$ is denoted by $\partial \mathcal{S}$; $\mathbf{I}$ denotes identity matrix of appropriate dimensions.

\section{Problem Formulation}\label{sec:prob_form}
\subsection{Dynamics of quadrotor UAV}
\begin{figure}[!h]
	\centering
	\includegraphics[width=0.42\textwidth]{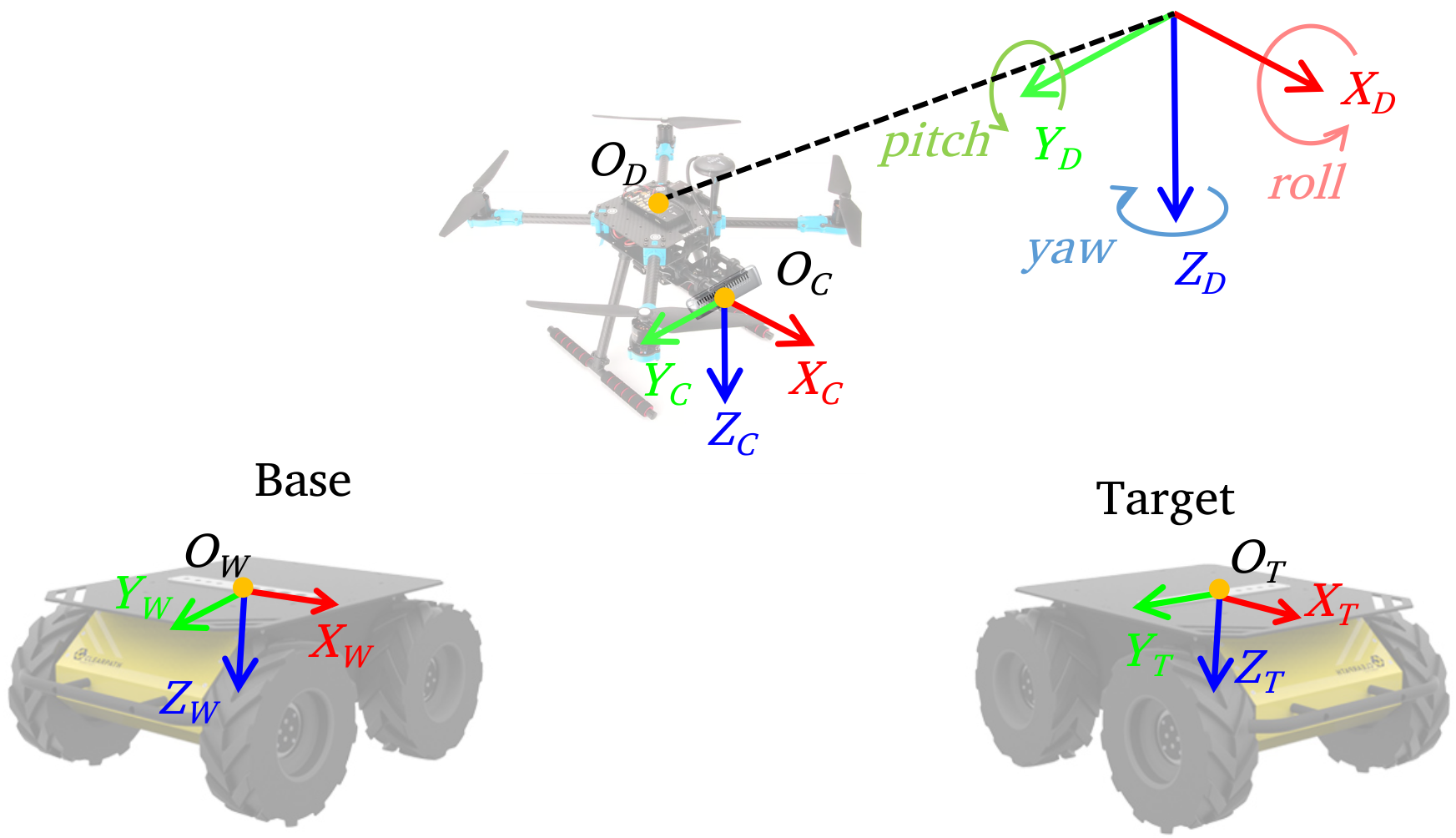}
	\caption{{A representation of the reference frames with $\mathbf{O}_D, \mathbf{O}_C, \mathbf{O}_W, \mathbf{O}_T$ as the origins of the UAV's body frame, camera frame, base frame, and target frame respectively. Their respective axes are subscripted with D, C, W, and T. }}\label{fig:frames}
\end{figure}
The UAV is modeled in the base frame $W$ and target frame $T$ (cf. Fig. \ref{fig:frames}) as given below, 
\begin{align}
    m \mathbf{\ddot{p}}_i(t) + m \mathbf{G} &= \boldsymbol{\tau}_{\mathbf{p}_i} - \mathbf{d}_{\mathbf{p}i}, \label{eq:p_tau} \\
    \mathbf{J}_i \mathbf{\ddot{q}}_i(t) + \mathbf{C}_i(\mathbf{q}_i, \mathbf{\dot{q}}_i, t)\mathbf{\dot{q}}_i (t) &= \boldsymbol{\tau}_{\mathbf{q}i} - \mathbf{d_{qi}}, \label{eq:q_tau} \\
		\boldsymbol{\tau}_{\mathbf{p}i} &= \mathbf{R}_i^D \mathbf{F} \label{eq:force_map}
\end{align}
where $\mathbf{p}_i \triangleq [x_i(t), y_i(t), z_i(t)]^T \in \mathbb{R}^3, \mathbf{q}_i \triangleq [\phi_i(t), \vartheta_i(t), \psi_i(t)]^T$ are the position and orientation of the UAV's center of mass $\mathbf{O}_D$ represented in the base and target frames respectively, where $i \in \lbrace W, T \rbrace$ represents the base and target frames, $m, \mathbf{J}_i \in \mathbb{R}^{3 \times 3}$ represent the UAV's mass and inertia matrices, $\mathbf{C}_i\in \mathbb{R}^{3 \times 3}$ represent the Coriolis or cross-coupling terms, $\mathbf{d}_{\mathbf{p}i}, \mathbf{d}_{\mathbf{q}i} \in \mathbb{R}^3$ represent the disturbances in the position and attitude dynamics, and $\boldsymbol{\tau}_{\mathbf{p}i}, \boldsymbol{\tau}_{\mathbf{q}i} \in \mathbb{R}^3$ represent the linear and angular control inputs in the base and target frames respectively, $\mathbf{G} \triangleq [0, 0, -9.81]^T\in \mathbb{R}^3$ is the gravitational vector, $\mathbf{F}\in \mathbb{R}^3$ is the thrust vector in the body fixed frame $\mathbf{X}_D - \mathbf{Y}_D - \mathbf{Z}_D$, mapped to the base and target frames using the respective rotation matrices $\mathbf{R}_i^D \in \mathbb{R}^{3 \times 3}$. For the simplification of the problem, the camera frame axes are assumed to be parallel to that of the body frame axes, such that,
\begin{align}
    \mathbf{p}_D^C = - \mathbf{p}_C^D, \quad \mathbf{p}_D^i = \mathbf{p}_D^C + \mathbf{p}_C^i, \label{eq:body_camera}
\end{align}
where $\mathbf{p}_C^i, \mathbf{p}_D^i \in \mathbb{R}^3$ are the relative positions of $i^{\text{th}}$ frame ($i \in \lbrace W, T \rbrace$) represented in the camera frame and drone's body frame respectively. 
\subsection{Visual-locking constraint for ascending}
\begin{figure*}[!h]
	\centering
	\includegraphics[width=0.7\textwidth]{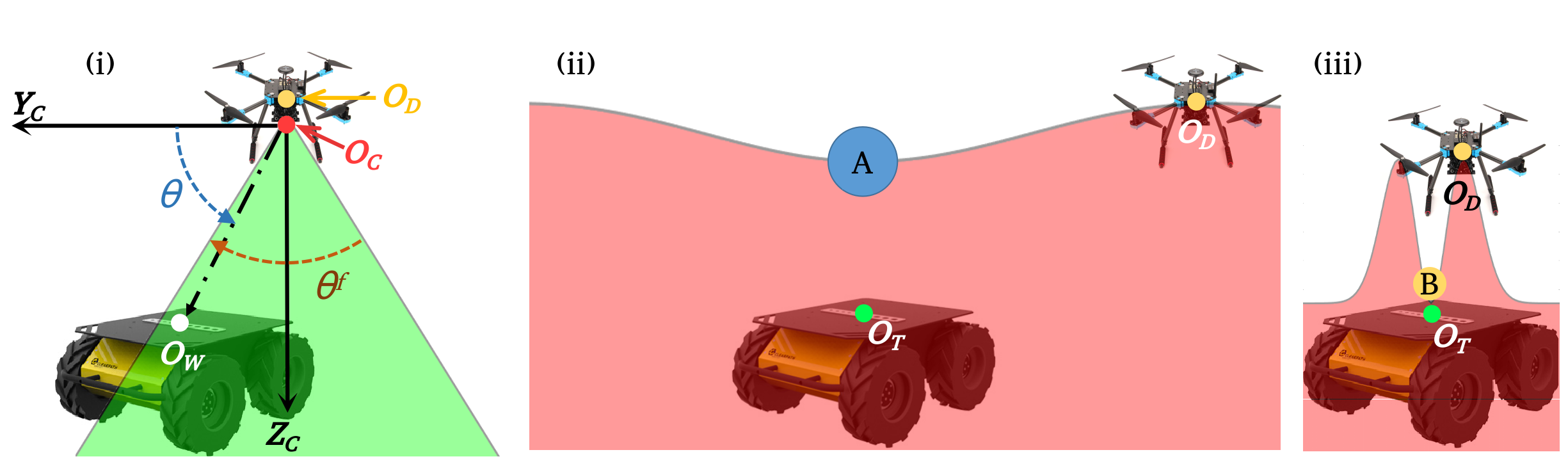}
	\caption{{A schematic representation of the constraints with respect to the UAV and UGVs: (i) the visual locking constraint between the camera frame and the base frame used in the ascending phase, where the green area is the safe region; (ii) the descending constraint used in the approaching phase between the UAV and the target, where the red area is an unsafe region, and region $A$ is the focus region; (iii) the descending constraint used in the landing phase, where region $B$ is the landing region. It is to be noted that constraints in (ii) and (iii) are formed with a single function with different parameters.}}\label{fig:concept}
\end{figure*}
In the ascending phase, the only available knowledge is the potential direction in which the target is placed with respect to the base frame. Thus, the following assumptions are made before formulating the constraint.
\begin{assumption}[Stationary base and target] \label{as:stationary}
    The base and target robots are stationary.
\end{assumption}

\begin{remark}
    Assumption \ref{as:stationary} is standard since in a coordinated task space, the robots would remain stationary during the switching operation.  
\end{remark}

\begin{assumption}[Proximity of the target] \label{as:proximity}
    The target is located at the proximity of the base such that there exists a region from which the UAV can track the features of both the base and the target.
\end{assumption}

\begin{remark}[Proximity]
    Assumption \ref{as:proximity} ensures the switching of the controller from one frame to another. Without the assumption, it is impossible to navigate the UAV between two robots since a global positioning is unavailable.
\end{remark}

The objective of the ascending phase is to travel along the direction of the target to detect it. During this maneuver, the localization of the UAV is based on the base target. So, the UAV must maintain the features of the base within its field of view, which results in the following conical constraint.
\begin{align}
    \text{tan}^{-1} \left (\frac{z_C^W}{\sqrt{l_C^W}} \right) > \frac{\pi}{2} - \frac{\theta^f}{2} , 
\end{align}
where $l_C^W = (x_C^W)^2 + (y_C^W)^2 \in \mathbb{R}^+$ is the squared horizontal distance between the UAV and the base in the camera frame, $\theta^f$ is the constrained field of view in radians (cf. Fig. \ref{fig:concept} (i)). The constraint transforms to the body frame as,
\begin{align}
    \text{tan}^{-1} \left(\frac{z_D^W - z_D^C}{\sqrt{l_D^W}} \right) > \frac{\pi}{2} - \frac{\theta^f}{2} \label{eq:vcbf_const}, 
\end{align}
where $l_D^W \triangleq (x_D^W - x_D^C)^2 + (y_D^W - y_D^C)^2$ is the transformed squared horizontal distance.

\subsection{Descending constraint}
In a UAV landing scenario, it is always preferable to align the UAV vertically over the target and then descend towards the target. Such behavior ensures that the UAV does not crash into the landing platform due to ground effects. Hence, this phase is divided into two subphases: approaching and landing. In the approaching phase, the UAV moves to a focus region (cf. Fig. \ref{fig:concept} (ii)) vertically above the target, which will ensure that the UAV is at a sufficient altitude to track the features reliably, which improves the tracking and ensures that the features are within the FOV while realigning the heading. The relative altitude for the focus region is decided based on where the feature tracking performance is more reliable. When the UAV is inside the focus region and the heading is aligned with the target, the landing phase is activated, in which the UAV vertically descends until the landing region (cf. Fig. \ref{fig:concept} (iii)), below which the feature tracking is unreliable, and the motors are turned off gradually to complete the landing.

A single parameterizable descending constraint is used for both the subphases (cf. Fig. \ref{fig:concept} (ii), (iii)), as given below:
\begin{align}
    z_T^D <  - K_1 K_2 (l_T^D) \text{exp}(-K_1 l_T^D) - K_3, \label{eq:dcbf_const} 
\end{align}
where $l_T^D \triangleq (x_T^D)^2 + (y_T^D)^2 \in \mathbb{R}^+$ is the squared horizontal distance between the UAV and the target in the target frame, and the parameters $K_1, K_2, K_3 \in \mathbb{R}^+$ are used to scale the CBF horizontally, scale it vertically, and shift it vertically respectively. $K_3$ is chosen to be the altitude of the focus region in the approaching phase, and the landing region in the landing phase. For a smooth transition between the CBF boundaries, the peak height of the DCBF boundary (${z_T^D}^*$) is chosen to be the altitude of the UAV when the switching happens. Subsequently, $K_1, K_2$ are obtained by finding the peak of the $h_d$, 
\begin{align}
    \frac{\partial z_T^D }{\partial l_T^D} &= - K_1 K_2 (1 - K_1 l_T^D)\text{exp}(-K_1 l_T^D), \nonumber \\
    \frac{\partial z_T^D }{\partial l_T^D} &= 0 \implies K_1 = \frac{1}{l_T^D} \implies 
    K_2 = -2.718({z_T^D}^* + K_3), \label{eq:K1_K2}
\end{align}
where ${z_T^D}^*$ is the relative altitude of the UAV in the target frame when the switching happens. If ${z_T^D} > -K_3$ when the target is detected, ${z_T^D}^*$ is chosen as ${z_T^D}^* = - K_3, \implies K_2 = 0$. Therefore, by choosing the altitudes of the focus region and landing region, the parameters can be found from the relative position of the UAV in the target frame using \eqref{eq:K1_K2} during the switching.

\section{Control Architecture} \label{sec:con_arc}
\begin{figure}[!h]
	\centering
	\includegraphics[width=0.4\textwidth]{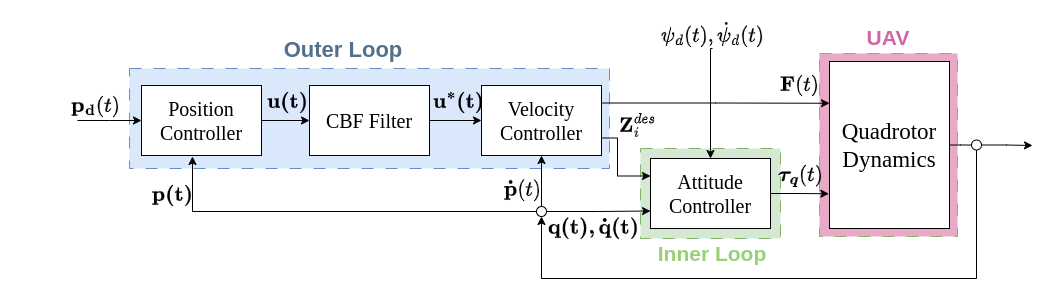}
	\caption{{Block diagram of the proposed control architecture.}}\label{fig:arch}
\end{figure}

The control architecture has a dual-loop structure, as the quadrotor UAV's linear and attitude dynamics are partly decoupled (cf. Fig. \ref{fig:arch}). The position controller is handled by the outer loop that consists of a) a nominal velocity input generator, b) a CBF layer to filter the velocity input, and c) an adaptive velocity tracking controller. 

Since the constraints are designed with respect to the body frame in the ascending phase, the pseudo virtual velocity control input $\mathbf{u}_D(t)$ is chosen in the body frame to be $1$ m/s along the direction of the target. In the descending phase, it is chosen as $\mathbf{u}_T(t) = \mathbf{K} \mathbf{p}_T^D(t)$ in the target frame as the constraints are in the target frame, where $\mathbf{K} \in \mathbb{R}^{3 \times 3}$ is a positive definite gain matrix. 

The CBF layer filters the pseudo virtual velocity input to accommodate the appropriate constraint and produces the corresponding virtual velocity input, $\mathbf{u}_D^*, \mathbf{u}_T^*$. The implementation of CBF is described later in section \ref{sec:cbf_design}. To compensate for the parametric uncertainties and external disturbances, an adaptive controller is implemented using the following steps,
\begin{align}
    \mathbf{e} &= \mathbf{\dot{p}}_i - \mathbf{\dot{p}}_i^{des}, \label{eq:error}
\end{align}
where the desired velocities for the phases are calculated as,
\begin{align}
    \mathbf{\dot{p}}_W^{des} = \mathbf{R}_W^D \mathbf{u}_D^*, \qquad \mathbf{\dot{p}}_T^{des} = \mathbf{u}_T^*, \label{eq:des_vel}
\end{align}
 Multiplying the time derivative of \eqref{eq:error} with $m$ and using \eqref{eq:p_tau}, we have
\begin{align}
    m \mathbf{\dot{e}} &= m(\mathbf{\ddot{p}}_i - \mathbf{\ddot{p}}^{des}) = \boldsymbol{\tau}_{\mathbf{p}i} - m\mathbf{G} + \varphi,
\end{align}
where $\varphi \triangleq -(m\mathbf{\ddot{p}}_i^{des} + \mathbf{d}_{\mathbf{p}_i})$ is the overall uncertainty, which can be upper bounded as,
\begin{align}
    ||\varphi|| \leq \kappa
\end{align}
The control law for the outer loop is designed as,
\begin{align}
    \boldsymbol{\tau}_{\mathbf{p}i}(t) &= - \mathbf{K_v} \mathbf{e}(t) - \widehat{\kappa}(t) \frac{e}{||e||} + \widehat{m}(t)\mathbf{G} \label{eq:pos_control}
\end{align}
where $\mathbf{K_v} \in \mathbb{R}^{3 \times 3}$ is a positive definite gain matrix, and the adaptive laws for the gains $\widehat{\kappa}, \widehat{m}$ are given by,
\begin{align}
    \dot{\widehat{\kappa}} &= ||\mathbf{e}|| - \eta_{\kappa} \widehat{\kappa}(t), ~\widehat{\kappa}(0) > 0 \label{eq:kappa} \\
    \dot{\widehat{m}} &= - \mathbf{e}^T\mathbf{G} - \eta_m \widehat{m}(t), ~\widehat{m}(0) > 0, \label{eq:hatM}
\end{align}
where $\eta_{\kappa}, \eta_{m}$ are saturation parameters. The closed-loop stability of the system and the convergence of error trajectory can be verified similarly to the proof in \cite{sankaranarayanan2020aerial}. $\boldsymbol{\tau}_{\mathbf{p}_i}$ is converted to the body frame using the relationship in \eqref{eq:force_map}.

The inputs to the inner loop control are generated by forming the desired rotation matrix, $\mathbf{R}_i^{des} \triangleq \begin{bmatrix} \mathbf{X}_i^{des} & \mathbf{Y}_i^{des} & \mathbf{Z}_i^{des} \end{bmatrix} \in \mathbb{R}^{3 \times 3}$, where the column matrices are given by,
\begin{align}
    \mathbf{Z}_i^{des} &= - \frac{\boldsymbol{\tau}_{\mathbf{p}i}}{||\boldsymbol{\tau}_{\mathbf{p}i}||}, 
    &\mathbf{X}_i^{des'} &= \begin{bmatrix}
        1 & 0 & 0
    \end{bmatrix}^T, \\    
    \mathbf{Y}_i^{des} &= \frac{\mathbf{Z}_i^{des} \times \mathbf{X}_i^{des'}}{||\mathbf{Z}_i^{des} \times \mathbf{X}_i^{des'}||}, 
    &\mathbf{X}_i^{des} &= \mathbf{Y}_i^{des} \times \mathbf{Z}_i^{des},
\end{align}

A PID controller is used to track the error in the orientation to track the Euler angles $\mathbf{q}_i^{des}$ obtained by converting the rotation matrix $\mathbf{R}_i^{des}$,
\begin{align}
    \boldsymbol{\tau}_{\mathbf{q}} &= - \mathbf{K}_{p}\boldsymbol{\epsilon} - \mathbf{K}_{d}\boldsymbol{\dot{\epsilon}} - \mathbf{K}_{i} \int \boldsymbol{\epsilon} dt, \label{eq:att_control}
\end{align}
where, $\boldsymbol{\epsilon} \triangleq \frac{1}{2}((\mathbf{R}_W^D)^T \mathbf{R}_i^{des} - (\mathbf{R}_i^{des})^T \mathbf{R}_W^D)^v$, $\dot{\epsilon} = (\mathbf{R}_W^D)^T \mathbf{R}_i^{des}\mathbf{\dot{q}}_{des} - \mathbf{\dot{q}}_i $ are the attitude and angular velocity tracking errors, $\mathbf{K}_p, \mathbf{K}_d, \mathbf{K}_i \in \mathbb{R}^{3 \times 3}$ are positive definite gain matrices. The thrust obtained from \eqref{eq:pos_control} is converted to body frame using the relationship \eqref{eq:force_map}, and used along with the inner loop control law \eqref{eq:att_control} to generate the necessary motor inputs.

\section{Control Barrier Function Design} \label{sec:cbf_design}
The constraints presented in \eqref{eq:vcbf_const} and \eqref{eq:dcbf_const} are nonlinear in nature, which requires a long prediction horizon to be implemented with a conventional model predictive control. It not only increases the performance demand but also introduces nonconvexity in the optimization problem, whose effects are well-known. Hence, we use the control barrier function, which ensures safety guarantees by rendering its super-level safe set forward invariant and asymptotically stable. For an affine function $\dot{x} = f(x) + g(x)u$, where $x \in \mathcal{X} \subset \mathbb{R}^n, u\in \mathcal{U} \subset \mathbb{R}^m $, where $f, g$ are Lipschitz continuous functions, a set representing the safe region of operation $\mathcal{S} \subset \mathcal{X}$ is rendered safe, if the control input $u$ ensures positive invariance of the set, i.e., $x(t_0) \in \mathcal{S} => x(t) \in \mathcal{S} ~\forall t \geq t_0$. Further, a measure of robustness can be incorporated into the notion of safety, if $\mathcal{S}$ is asymptotically stable when initialized in a set $\mathcal{D} \setminus \mathcal{S}$, where $\mathcal{S} \subset \mathcal{D} \subset \mathcal{X}$. 

\begin{definition} \label{def:cbf}
    A continuously differentiable function $h(x): \mathcal{D} \rightarrow \mathbb{R}$ with the safe set $\mathcal{S}$ as a zero super-level set of $h(x)$, i.e., $\mathcal{S}:= \lbrace x \in \mathcal{X} \mid h(x) \geq 0 \rbrace$, is a control barrier function, if there exists a real parameter $\gamma > 0$ and a generalized class-$\mathcal{K}$ function $\alpha$, such that for all $x \in \mathcal{D}$,
    \begin{align}
        \underset{u \in \mathcal{U}}{sup} \left \lbrace L_f h(x) + L_g h(x) u + \gamma \alpha(h(x)) \right \rbrace \geq 0. \label{eq:cbf_def}
    \end{align}
\end{definition}

The forward invariance of $\mathcal{S}$ ($\dot{h} \geq 0$ on $\partial \mathcal{S}$) and asymptotically stability of $\mathcal{S}$ ($\dot{h} \geq 0$ on $\mathcal{D} \setminus \mathcal{S}$) are captured together in condition \eqref{eq:cbf_def}. The CBF design for the different phases is described in the following subsections.

\subsection{Visual locking CBF}
The position kinematics of the UAV in the body frame can be approximated to,
\begin{align}
    \mathbf{\dot{p}}^W_D = \mathbf{u}_1, \label{eq:as_pos_kin}
\end{align}
provided the lower-level controllers track their inputs at a faster rate with accuracy. From the visual locking constraint provided in \eqref{eq:vcbf_const}, the VCBF constraint is designed as,
\begin{align}
    h_v = \text{tan}^{-1} \left (\frac{z_D^W - z_D^C}{\sqrt{l_D^W}} \right) - \frac{\pi}{2} + \frac{\theta^f}{2} . \label{eq:vcbf}
\end{align}
The partial derivatives of \eqref{eq:vcbf} are given by,
\begin{subequations} \label{eq:hv_derivative}
\begin{align}
\frac{\partial h_v}{\partial \mathbf{p}_D^W} &=
\begin{bmatrix}
    \frac{\partial h_v}{\partial x_D^W} \\[10pt] \frac{\partial h_v}{\partial y_D^W} \\[10pt] \frac{\partial h_v}{\partial z_D^W}
\end{bmatrix} & = \begin{bmatrix}
    -\frac{(x_D^W - x_D^C)(z_D^W - z_D^C)}{((z_D^W - z_D^C)^2 + l_D^W)\sqrt{l_D^W}}  \\[10pt] -\frac{(y_D^W - y_D^C)(z_D^W - z_D^C)}{((z_D^W - z_D^C)^2 + l_D^W)\sqrt{l_D^W}} \\[10pt] \frac{\sqrt{l_D^W}}{(z_D^W - z_D^C)^2 + l_D^W}
\end{bmatrix} 
\end{align}
\end{subequations}
It can be shown that the candidate barrier function in \eqref{eq:vcbf}, satisfies the Definition \ref{def:cbf} for the admissible control set $\mathcal{U} = [-v_m^x, v_m^x] \times [-v_m^y, v_m^y] \times [-v_m^z, v_m^z]$, where $v_m^x, v_m^y, v_m^z \in \mathbb{R}^+$ define the absolute bounds on the velocities of the UAV in the body frame. So, $h_v$ is proven to be a valid control barrier function for the system \eqref{eq:as_pos_kin} in accordance with \cite{ames2019control}. Therefore, the relative position of the base $\mathbf{p}_{D}^W \in \mathbb{R}^3$ remains within the visual constraints established by the zero super level set $\mathcal{S}_1 = \lbrace \mathbf{p}_{D}^W \in \mathbb{R}^3 \mid h_v(\mathbf{p}_{D}^W) \geq 0 \rbrace$ as formulated in \eqref{eq:vcbf}. The filtered virtual velocity input $\mathbf{u}_D^*$ in the body frame is obtained by optimizing the quadratic program that yields a control input, which is minimally deviating from the  pseudo virtual velocity input $\mathbf{u}_D$ while enforcing the safety constraint in \eqref{eq:vcbf_const} as shown below,
\begin{align}
    \mathbf{u}_D^*\left(\mathbf{p}_D^W(t) \right) &= \underset{\mathbf{u}_1 \in \mathcal{U}}{\text{argmin}} || \mathbf{u}_1 - \mathbf{u}_D || \nonumber \\
    s.t. &: \frac{\partial h_v}{\partial \mathbf{p}_D^W}\mathbf{u}_1 \geq - \alpha_v \left( h_v \left (\mathbf{p}_D^W \right) \right) \label{eq:vcbf_quad}
\end{align}
where 
    the constraint in \eqref{eq:vcbf_quad} is derived using \eqref{eq:cbf_def} and \eqref{eq:as_pos_kin} ($f(x) = 0, g(x) = \mathbf{I}$). It is evident that for a given $\mathbf{p}_D^W$, the constraint in \eqref{eq:vcbf_quad} is linear in $\mathbf{u}_1$, with the coefficients defined by the partial derivatives in \eqref{eq:hv_derivative}, which can be solved at a high speed. It is to be noted that though $h_v$ is discontinuous at $x_C^W = y_C^W = 0.0$, the UAV would not be in that position as $\mathbf{u}_D^T$ is always pointing away from the origin of the base.

\subsection{Descending CBF}
Similar to \eqref{eq:as_pos_kin}, the position kinematics of the UAV is represented in the target frame as, 
\begin{align}
    \mathbf{\dot{p}}_T = \mathbf{u}_2, \label{eq:de_pos_kin}
\end{align}
From \eqref{eq:dcbf_const}, the CBF constraint is designed as,
\begin{align}
    h_d  = - z_T^D  - K_1 K_2 (l_T^D) \text{exp}(-K_1 l_T^D) - K_3, \label{eq:dcbf} 
\end{align}
The partial derivatives of \eqref{eq:dcbf} is given by,
\begin{align}
    \frac{\partial h_d}{\partial \mathbf{p}_T} =
    \begin{bmatrix}
        \frac{\partial h_d}{\partial x_T} \\ \frac{\partial h_d}{\partial y_T} \\ \frac{\partial h_d}{\partial z_T}
    \end{bmatrix} &= \begin{bmatrix} 2 K_1 K_2 x_T (K_1 l_T^D - 1) \text{exp}(-K_1 l_T^D) \\ 2 K_1 K_2 y_T (K_1 l_T^D - 1) \text{exp}(-K_1 l_T^D) \\ -1 \end{bmatrix} \label{eq:hd_derivative}
\end{align}
By the definition of the CBF in \ref{def:cbf}, it can be shown that the candidate CBF presented in \eqref{eq:dcbf} is valid for the admissible control set $\mathcal{U} = [-v_m^x, v_m^x] \times [-v_m^y, v_m^y] \times [-v_m^z, v_m^z]$ where $v_m^x, v_m^y, v_m^z \in \mathbb{R}$ define the absolute bounds on the velocities of the UAV in the target frame. So, $h_d$ is proven to be a valid control barrier function for the system \eqref{eq:de_pos_kin} in accordance with \cite{ames2019control}. Therefore, the UAV's position in the target frame $\mathbf{p}_{T} \in \mathbb{R}^3$ remains within the descending constraints established by the zero super level set $\mathcal{S}_2 = \lbrace \mathbf{p}_{T} \in \mathbb{R}^3 \mid h_d(\mathbf{p}_{T}) \geq 0 \rbrace$ as formulated in \eqref{eq:dcbf}. The following quadratic program is formed to obtain a filtered virtual velocity input $\mathbf{u}^*_T$, which enforces the safety constraint in \eqref{eq:dcbf_const}, with only a minimal deviation from the pseudo virtual velocity input $\mathbf{u}_T$,
\begin{align}
    \mathbf{u}_T^*\left(\mathbf{p}_T(t) \right) &= \underset{\mathbf{u}_2 \in \mathcal{U}}{\text{argmin}} || \mathbf{u}_2 - \mathbf{u}_T || \nonumber \\
    s.t. &: \frac{\partial h_d}{\partial \mathbf{p}_T}\mathbf{u}_2 \geq - \alpha_p \left( h_d \left (\mathbf{p}_T \right) \right) \label{eq:dcbf_quad}
\end{align}
where the constraint in \eqref{eq:dcbf_quad} is derived using \eqref{eq:cbf_def} and \eqref{eq:de_pos_kin} ($f(x) = 0, g(x) = \mathbf{I}$). It is evident that for a given $\mathbf{p}_T$, the constraint in \eqref{eq:dcbf_quad} is linear in $\mathbf{u}_2$, with the coefficients defined by the partial derivatives in \eqref{eq:hd_derivative}, which can be solved at a high speed. It is to be noted that though $h_d$ is discontinuous at $l_T^D = 0.0$, the UAV would be perfectly aligned with the target, and the landing can be performed at that instant with the nominal controller alone.

\section{Experimental Validation} \label{sec:exp_val}
The control architecture is experimentally validated using a Holybro X500 quadrotor UAV ascending from a Husky UGV and landing on another. To validate the precise landing performance, a landing platform is attached to the target UGV, whose margin of error is $\pm 0.02$ m along the horizontal plane. The UAV is equipped with an onboard computer Lattepanda 3 Delta running ROS2 on a Linux environment. The attitude control is handled using an onboard flight controller equipped with an IMU. A 3x3 array of 6x6\_250 ArUco markers (length $= \unit[0.13]{m}$) is placed in the base and a single 4x4\_250 ArUco marker is placed on the target. The ArUco detection was performed using the OpenCV ArUco library with an Intel RealSense D455 camera fixed running at $30$ FPS. To test the robustness of the control architecture, the UAV's pose is estimated naively from the feature tracking without fusing the IMU information, and the velocities are estimated from the IMU. The matrices
\begin{align*}
    \mathbf{T}_6^W = \begin{bmatrix}
        0 & -1 & 0 & 0.23 \\
        1 & 0 & 0 & 0.33 \\
        0 & 0 & 1 & 0 \\
        0 & 0 & 0 & 1
    \end{bmatrix}, & 
    \mathbf{T}_4^T = \begin{bmatrix}
        0 & 1 & 0 & 0.06 \\
        -1 & 0 & 0 & -0.07 \\
        0 & 0 & 1 & 0 \\
        0 & 0 & 0 & 1
    \end{bmatrix},
\end{align*}
represent the transformations between the 6x6 ArUco frame and base frame, and 4x4 ArUco frame and target frame respectively. The parameters used in the experiment are $\mathbf{K} = 1.2\mathbf{I}, \widehat{\kappa}(0) = 0.01, \widehat{m}(0) = 0.1, \eta_{\kappa} = 2.5, \eta_m = 0.5, \alpha_v = 5, \alpha_d = 3.5, \mathbf{K}_p = 0.1\mathbf{I}, \mathbf{K}_d = 0.03\mathbf{I}, \mathbf{K}_i = 0.01\mathbf{I}, \theta^f = 50^0$, and the relative distance between the camera frame and drone frame is given by $\mathbf{p}_D^C = [-0.1, 0.0, 0.1]^T\unit{m}$. For the given size of the marker and the camera resolution, the marker detection is reliable between $\unit[0.35]{m}$ to $\unit[1.75]{m}$ altitude. So, they are chosen as the $K_3$ values at the landing phase and approaching phase respectively. The switching to the landing phase happens when the UAV enters the ball with a radius of 0.1 m around the focus point and the yaw is aligned within 5 degrees of error. The upper bounds of the filtered control input are set to be $v_m^x = v_m^y = v_m^z = \unit[0.1]{m/s}$. Two runs of the experiment are carried out with two different settings. The rotations between the base and target frames in the settings $\mathbf{R}_W^{T1}, \mathbf{R}_W^{T2}$ are given by,
\begin{align*}
    \mathbf{R}_W^{T1} & = \begin{bmatrix}
        0 & 1 & 0  \\ -1 & 0 & 0  \\ 0 & 0 & 1  
    \end{bmatrix}, & \mathbf{R}_W^{T2}  &= \begin{bmatrix}
        -0.76 & -0.65 & 0  \\ 0.65 & -0.76 & 0  \\ 0 & 0 & 1  
    \end{bmatrix},
\end{align*}
their relative positions are given by $\mathbf{p}_W^{T1} \triangleq [1.1, -0.1, -0.07]^T\unit{m}$, $\mathbf{p}_W^{T2} \triangleq [-1, -0.5, -0.07]^T$\unit{m}, and the a priori velocities in the first and second settings are chosen to be $\mathbf{u}_D \triangleq [1, 0, 0]^T\unit{m/s}$ and $\mathbf{u}_D \triangleq [-1, 0, 0]^T\unit{m/s}$ respectively. It is to be noted that the velocity inputs in both settings are not exactly toward the target but approximately in the direction of the target. A wind disturbance of about $\unit[5]{m/s}$ is added in both settings. The results and analysis of the experiments are presented in the following subsection.

\subsection{Results and Analysis}
\begin{figure}[!h]
	\centering
	\includegraphics[width=0.44\textwidth]{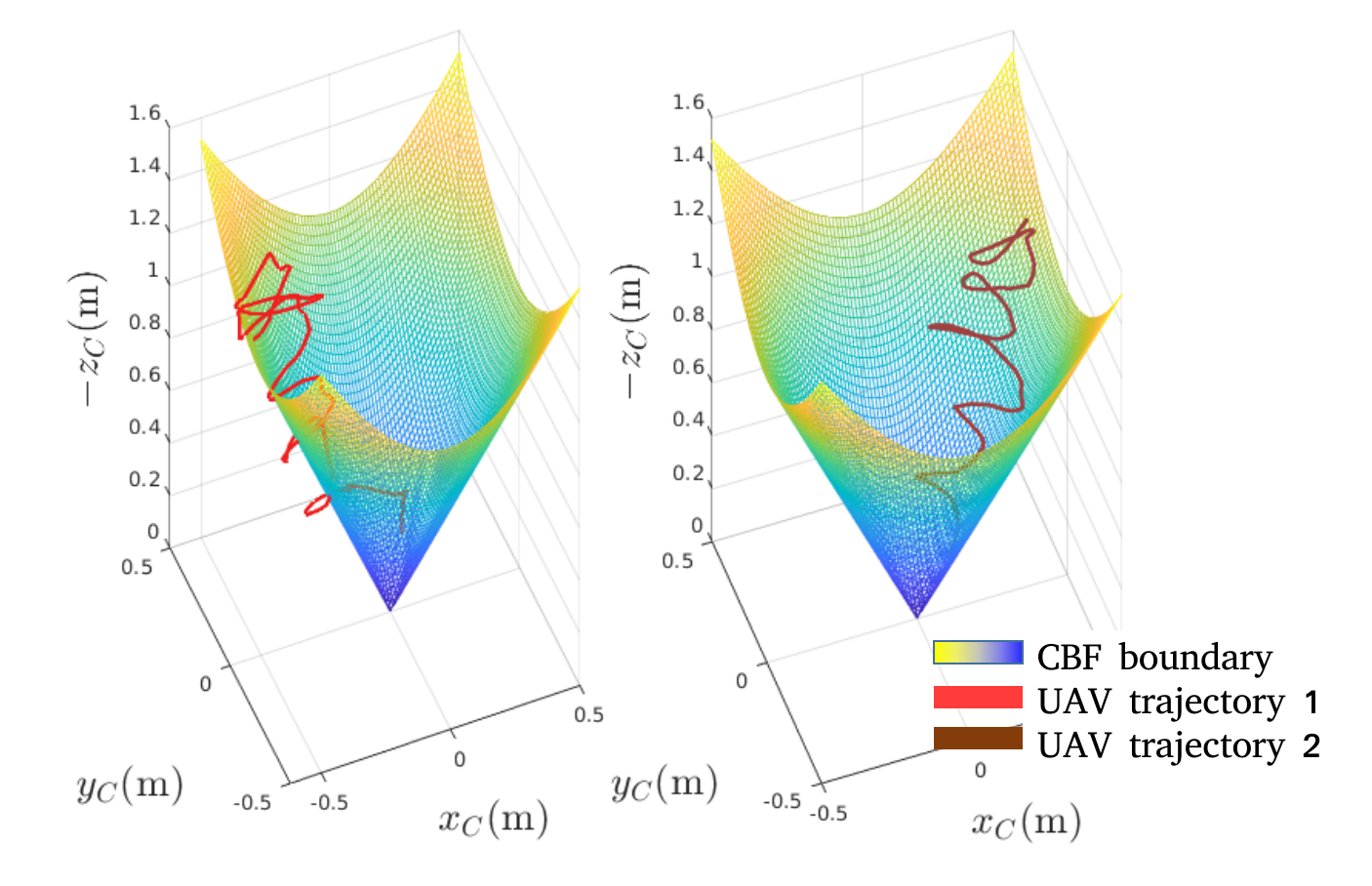}
	\caption{{The boundary layer of the VCBF used in the ascending phase with the relative trajectory of the base frame represented in the camera frame for both settings. }}\label{fig:ascending}
\end{figure}
\begin{figure}[!h]
	\centering
	\includegraphics[width=0.44\textwidth]{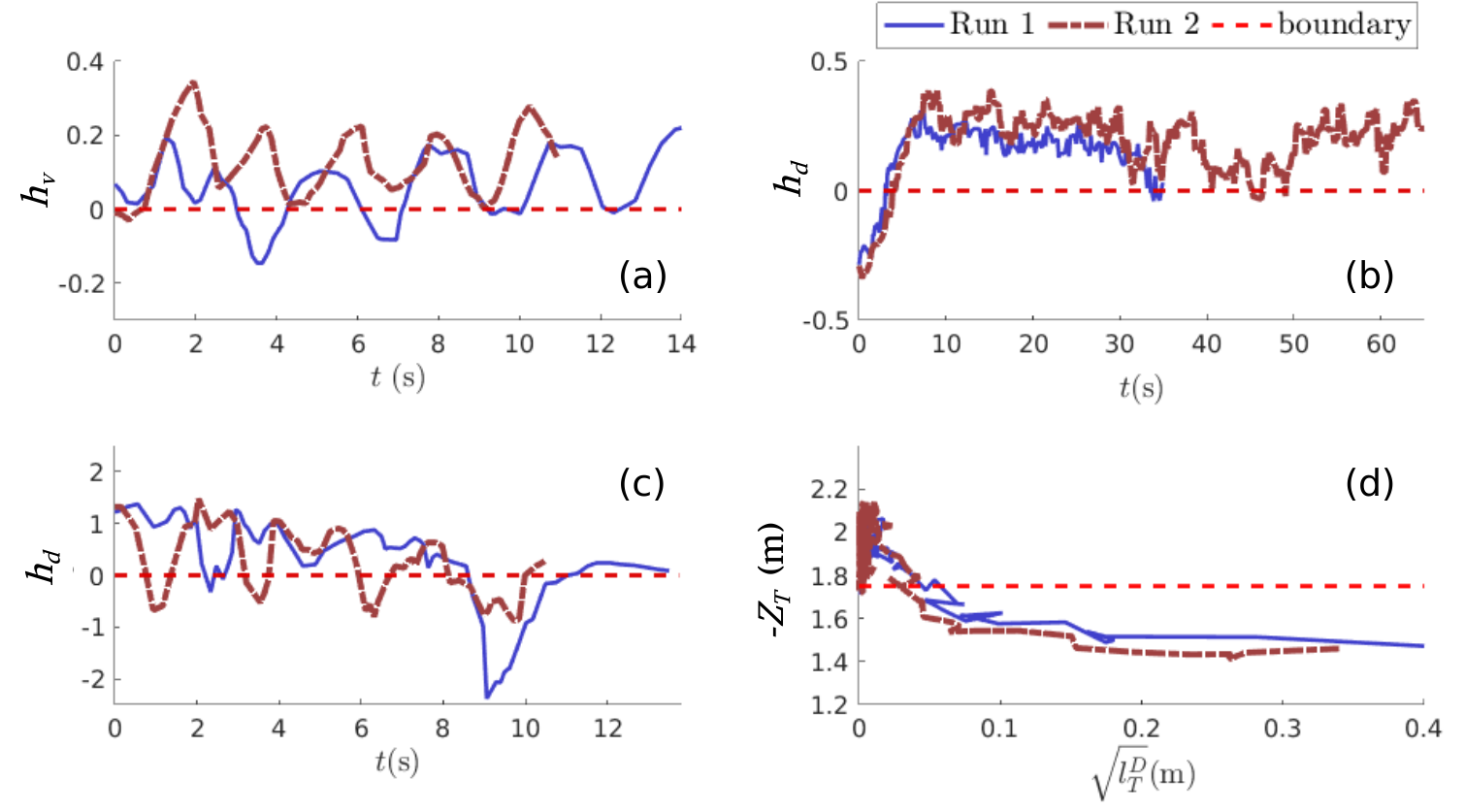}
	\caption{{The values of the CBF $h_v, h_d$ during (a) ascending, (b) approaching, and (c) descending phases for both the runs and (d) the plot between the vertical and horizontal distance of the UAV in the target frame with the boundaries.}}\label{fig:approach_h_values}
\end{figure}
\begin{figure}[!h]
	\centering
	\includegraphics[width=0.42\textwidth]{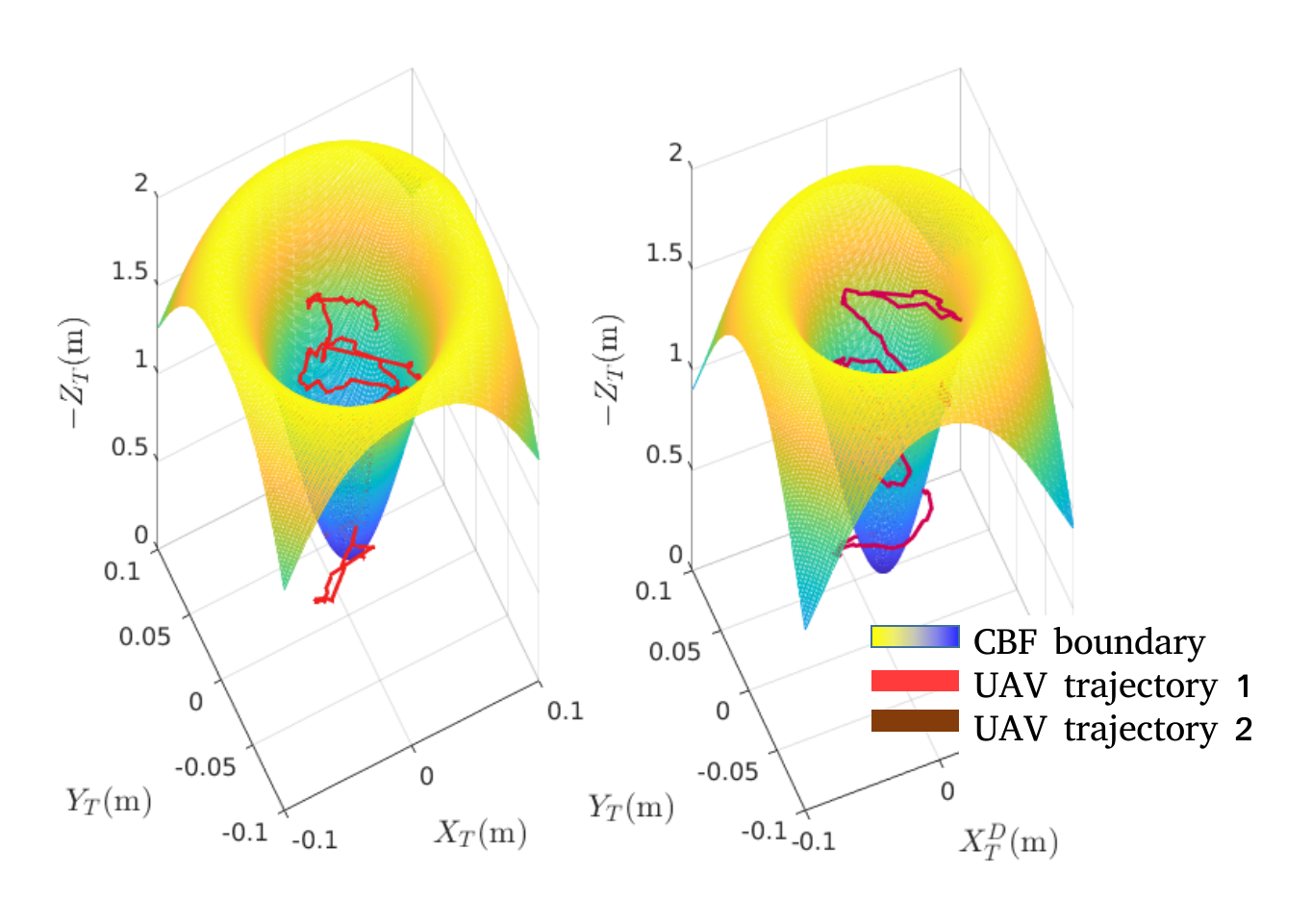}
	\caption{{The boundary layer of the DCBF used in the landing phase with the trajectory of the UAV in the target frame for both settings.}}\label{fig:landing}
\end{figure}
\begin{figure}[!h]
	\centering
	\includegraphics[width=0.44\textwidth]{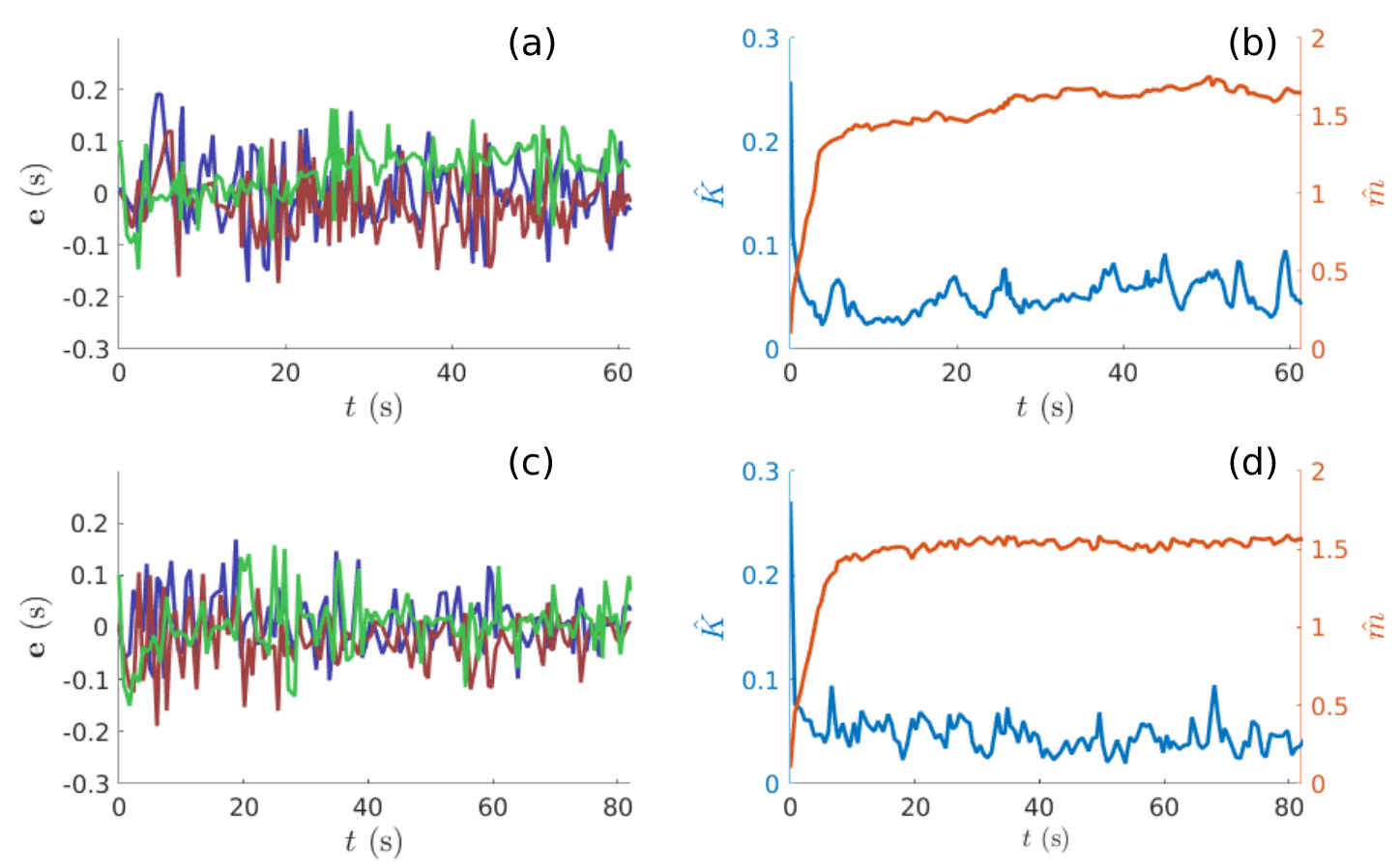}
	\caption{{(a) and (c) show the error trajectories of the adaptive velocity controller for the first run and second run respectively, where the blue, brown and green lines represent the velocity tracking errors in $x, y, z$ axes of the body frame. (b) and (d) show the adaptive gains $\widehat{\kappa}, \widehat{m}$ in the first and second runs respectively.}}\label{fig:adaptive}
\end{figure}
The target was detected at an altitude below the decided $K_3$ in both runs. So, the gain values calculated based on \eqref{eq:K1_K2} resulted in the following values, in the approaching phase $K_2 = 0$, $K_1 = 2.38$ in the first run, and $2.94$ in the second run. During the landing phase, $K_1 = 120.6, K_2 = 4.2$ in the first run, and $K_1 = 156.1, K_2 = 4.0$ in the second run. Figures~\ref{fig:ascending}-\ref{fig:adaptive} highlight the controller's performance. Figure~\ref{fig:ascending} shows the relative position of the base in the camera frame along with the conical boundary of the VCBF constraint. It is observed that inside the safe region, the CBF filter allows the UAV to move horizontally, whereas when it reaches the boundary, the filter pushes the UAV inward and upward. Since the constraints are chosen to be conservative compared to the actual FOV of the camera ($90^0 \times 65^0)$, the tracking algorithm does not lose the markers for small breaches of the constraint, which are observed in Run 1 due to disturbances and the upper bounds of the control inputs, which indirectly validates the robustness of the CBF's asymptotic convergence property. A similar pattern is observed on the landing constraint in Fig. \ref{fig:landing}. It is to be noted that the Z-axis is flipped in both graphs (Figs. \ref{fig:ascending} \& \ref{fig:landing}) for a vertical perspective of the CBFs' boundary layers and the trajectories. The DCBF forms a cone-like structure, which encloses the UAV's trajectories to remain within a minimal margin of error for precise landing.

Since, the gain $K_2 = 0$ in the approaching phase, the constraint becomes a plane perpendicular to the $\mathbf{Z}_T$ axis. Fig. \ref{fig:approach_h_values} (d) shows a plot between the horizontal and vertical distance of the UAV from the base along with the DCBF's boundary. The asymptotic convergence to the safe set initiating from an unsafe region is evident from the plot. During the transient phase, where the UAV is aligning its horizontal position and its heading with those of the target, the DCBF ensures that the constraints are enforced in the process, which causes an oscillation around the focus region, before switching to the landing phase. The values of $h_v, h_d$ functions in different phases are shown in Fig. \ref{fig:approach_h_values} (a), (b), (c). Fig. \ref{fig:adaptive} presents the tracking error and the variation of gains $\widehat{\kappa}, \widehat{m}$ over the experiment during the two runs. The controller effectively adapts the control parameters to tackle the uncertainties and the varying wind speed to aid the CBF layer in enforcing the constraints and reducing the deviations to ensure that the UAV does not lose the features or collide with the target. Even in the presence of uncertainties in the measurement and the dynamics, the combination of adaptive control and the CBFs guarantees safe navigation and landing in both runs.

\section{Conclusions} \label{sec:conc}
A switched CBF-based systematic control method is developed for the visual navigation of a quadrotor UAV between two ground robots localizing only using a downfacing vision sensor that tracks visual features placed on the surface robots. An adaptive controller is designed to track the velocity control inputs in the presence of modeling uncertainties and external disturbances. The controller performance is experimentally validated in two different settings with wind disturbance. In future work, we would extend the problem to a moving UGV scenario. Further, we would use a more robust visual locking constraint to accommodate the rectangular field of view of the camera in both ascending and descending phases to ensure that the UAV does not lose the target while changing its heading. 

\bibliographystyle{IEEEtran}
\bibliography{root}
\end{document}